\documentclass{article}
\usepackage{arxiv}
\usepackage[utf8]{inputenc} 
\usepackage[T1]{fontenc}    
\usepackage{hyperref}       
\usepackage{url}            
\usepackage{booktabs}       
\usepackage{amsfonts}       
\usepackage{nicefrac}       
\usepackage{microtype}      
\usepackage{lipsum}		    
\usepackage{graphicx}
\usepackage[numbers]{natbib}
\usepackage{doi}

\usepackage{amsmath}
\usepackage{makecell}
\usepackage{booktabs}
\usepackage[table]{xcolor}
\usepackage{caption}
\usepackage{listings}
\usepackage{xcolor}
\usepackage{pythonhighlight}

\title{On Linear Separability of the MNIST Handwritten Digits Dataset}

\date{November, 2025} 

\author{ \href{https://orcid.org/0000-0001-6410-8895}{\includegraphics[scale=0.06]{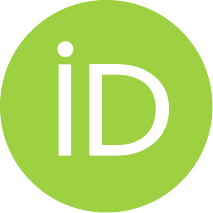}\hspace{1mm}\'Akos Hajnal}\textsuperscript{a,b}\\
	\textsuperscript{a}Laboratory of Parallel and Distributed Systems\\
        Institute for Computer Science and Control (SZTAKI) \\
        Hungarian Research Network (HUN-REN)\\
        \textsuperscript{b}John von Neumann Faculty of Informatics \\
        \'Obuda University \\
	Budapest, Hungary \\
	\texttt{akos.hajnal@sztaki.hu} \\
}

\renewcommand{\undertitle}{}
\renewcommand{\headeright}{}

\hypersetup{
pdftitle={On Linear Separability of the MNIST Handwritten Digits Dataset},
pdfauthor={Akos Hajnal},
pdfkeywords={MNIST dataset, Linear separability, Machine learning},
}

\begin{document}
\maketitle

\begin{abstract}
The MNIST dataset containing thousands of handwritten digit images is still a fundamental benchmark for evaluating various pattern-recognition and image-classification models. Linear separability is a key concept in many statistical and machine-learning techniques. Despite the long history of the MNIST dataset and its relative simplicity in size and resolution, the question of whether the dataset is linearly separable has never been fully answered -- scientific and informal sources share conflicting claims. This paper aims to provide a comprehensive empirical investigation to address this question, distinguishing pairwise and one-vs-rest separation of the training, the test and the combined sets, respectively. It reviews the theoretical approaches to assessing linear separability, alongside state-of-the-art methods and tools, then systematically examines all relevant assemblies, and reports the findings.
\end{abstract}


\setcitestyle{square}

\section{Introduction}
\begin{figure*}[t]
\centering
\includegraphics[width=0.8\linewidth]{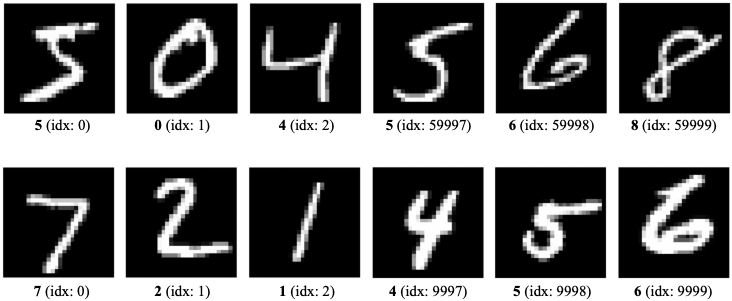}
\caption{Samples from the MNIST dataset. First and last three images of the training (first row) and the test set (second row) are shown. Sub-captions indicate the digit (label) and the index of the sample in the series.}\label{fig1}
\end{figure*}

The MNIST dataset created in the late 1990s by Yann LeCun at al. \cite{lecun1998a} is still a fundamental benchmark for evaluating pattern-recognition and image-classification methods. It consists of 70,000 gray-scale images of handwritten digits in 28x28-pixel resolution, as illustrated in Fig. \ref{fig1}. The dataset is split into 60,000 training and 10,000 test samples. The goal is to accurately recognize all digits in the test set by teaching the model exclusively on the training data.

Various machine learning approaches had been applied to this dataset, including linear models,  k-Nearest Neighbor, Random Forest, Support Vector Machines (SVM), both shallow and Deep Neural Networks, Convolutional Neural Networks (CNN). At the time of writing, the best published result was obtained using an ensemble of CNNs \cite{an2020ensemblesimpleconvolutionalneural}, achieving 99.91\% accuracy, which surpasses \textit{estimated} human performance of approximately 0.2\% error \cite{lecun1995}.

Linear models and the notion of linear separability, separating hyperplanes are central to many fields, including statistics and optimization. Classic algorithms like the Perceptron algorithm \cite{rosenblatt1957perceptron}, SVM \cite{cortes1995support}, logistic regression rely on these concepts, and constitute the fundamental basis of the operation performed by an artificial neuron.

Linear separability is the property of a dataset that allows a single linear decision boundary to separate the data points belonging to two classes. Formally, given \(n\) samples \(\{x_1,\dots,x_n\}\) (\(x_i \in \mathbb{R}^d\) for $i=1,..., n$, $d$ is the number of features) and labels \(y_i\in\{-1,1\}\) corresponding to one of the two classes associated with sample \(x_i\), the dataset is \textit{linearly separable} if there exist a separating hyperplane \(w^\top x_i + b = 0\) (\textit{normal vector} \(w \in \mathbb{R}^d\), \textit{bias} \(b\in \mathbb{R}\)) such that \(w^\top x_i + b > 0\) for all $x_i$ where $y_i=1$, otherwise \(w^\top x_i + b < 0\). The \textit{separating hyperplane}, if exists, therefore passes between the data points dividing all the points with positive labels on one side and negatives on the other side. \textit{Testing} linear separability is the task to only decide the existence of a separating hyperplane, without providing one.

Samples in the MNIST dataset fall into ten classes corresponding to the digits 0--9. Linear separability can be investigated in two settings: (1) \textit{pairwise linear separability} and (2) \textit{one-vs-rest linear separability}. In the pairwise case, we ask whether samples of one digit (e.g., 0) can be linearly separated from samples of another digit (e.g., 1). In the one-vs-rest case, we ask whether samples of a given digit (e.g., 0) can be linearly separated from all other digits (1--9).

Despite the long history of the MNIST dataset and its relative simplicity (in number of samples and resolution) only a few papers examined these questions thoroughly. Zhong et al. \cite{zhong2024} carried out experiments to determine linear separability of the MNIST \emph{test set} of digit 0 and each of the remaining 9 digits (pairwise), furthermore, computed the so called degree of linear separability for all digit pairs (45), but without claiming linear separability between these latter pairs. Another study \cite{e23030305} investigated a single digit pair (3 vs 7) on a reduced subset, on down-scaled samples, and stated that "the MNIST training set is not linearly separable" citing \cite{lecun1998a}, which paper did not actually make this claim. The authors in research \cite{monteiro2020binary} trained so called Phase Separation Binary Classifier models for each digit pair for the MNIST dataset, but they didn't address and report linear separability. \cite{crammer2010} conducted empirical study on all-pairs (45) and 1-vs-rest separation of the MNIST dataset; the results provided (accuracy comparison with another method) are however not conclusive with regard to the linear separability.

This paper aims to clarify the question of linear separability of the MNIST dataset and to provide comprehensive analysis and empirical evidence -- even if the findings might merely confirm the widespread but informal \emph{consensus} that "MNIST is \textit{not} linearly separable". Section \ref{sec:related} presents an overview of existing approaches and state-of-the-art methods. After selecting and applying a suitable technique, section \ref{sec:results} evaluates and reports the results of the linear separability experiments. Finally, the paper is concluded.

\section{Approaches to Testing Linear Separability}\label{sec:related}

Linear separability has been studied for many decades. Fisher's Linear Discriminant Analysis \cite{fisher1936} introduced a statistical method for separating datasets (1936), however, it does not directly address linear separability. The Perceptron algorithm \cite{rosenblatt1957perceptron} proposed by Rosenblatt (extended later by many variants, such as \cite{hajnal2023}) guarantees convergence at finding the separating hyperplane in finite steps -- provided the dataset is indeed separable. The number of steps required is typically unknown in advance (and can be extremely large); non-termination cannot conclude non-separability, regardless of the number of iterations performed so far. 

Support Vector Machine (SVM) introduced by Cortes and Vapnik \cite{cortes1995support} aimed to find the separating hyperplane with largest margin. Determining linear separability is a special case of support vector machines, known as a \textit{hard-margin SVM} with a linear kernel. SVM solvers, such as LIBSVM \cite{chang2011} (which is the solver backend in Scikit-learn and other data mining toolkits) include a cost parameter to penalize misclassifications. In practice, this parameter is set to a large value to enforce, or closely approximate perfect separation. However, when the dataset is high-dimensional or contains a large number of samples, this can demand an excessive number of iterations or yield non-perfect separation. 

Computational geometrical approaches treat data samples as points in a high-dimensional space. The separating hyperplane theorem guarantees linear separability, if the two classes of data form disjoint, convex sets. To verify this, one can compute the convex hulls of each class and then check whether these hulls intersect. The time complexity of computing the convex hulls (e.g. with the QuickHull algorithm \cite{barber1996quickhull}) can increase exponentially with data dimension.

Zhong et al. \cite{zhong2024} transformed the linear separability problem to an equivalent Minimum Enclosing (Covering) Ball (MEB) problem. Earlier MEB algorithms exhibited exponential time complexity \cite{kallberg2019minimum}; recent studies have come to a polynomial-time solution \cite{dearing2009dual, cavaleiro2018faster}. These findings make this approach competitive with linear programming solutions -- proved to be faster than Simplex or Interior-point methods in Matlab environment (Table VIII in \cite{zhong2024}).

A widely used approach to decide linear separability is to solve a linear program (LP) \cite{scholkopf1999advances}:

\begin{equation}\label{eqn1}
\begin{aligned}
& \min_{w,b} \quad 0 \\ 
& \ \text{subject to} \\
& \quad \quad \quad w^\top x_i + b \ge 1, & \forall i \ \text{with}\ y_i = +1 \\
& \quad \quad \quad w^\top x_j + b \le -1, & \forall j \ \text{with}\ y_j = -1
\end{aligned}
\end{equation}

or more concisely: 

\begin{equation}\label{eqn2}
\begin{aligned}
& \min_{w,b} 0 \quad \text{subject to} \quad y_i ( w^\top x_i + b ) \ge 1
\end{aligned}
\end{equation}

The objective function is a constant, so this formulation is purely a feasibility problem. If it is feasible, the solution for decision variables $w$ and $b$ represent the normal vector and the scalar bias of a separating hyperplane. Popular solvers include the Simplex method and the Interior-point method. While the Simplex methods exhibits exponential growth in time complexity (as the data dimension increases), interior-point method offers polynomial-time execution.

Rewriting the objective function to $min(\frac{1}{2}||w||^2)$ transforms the formulation into a quadratic program (QP), subject to convex optimization \cite{boyd2004convex}, providing the unique, \textit{optimal} separating hyperplane with maximum margin (out of the scope in this paper). Convex optimization guarantees global optimum and can also provide proof of infeasibility when no solution exists. CVXPY is an open-source tool for modeling and solving various convex optimization problems \cite{agrawal2018rewriting, diamond2016cvxpy}, which contains several state-of-the-art solver implementations. 

We chose CVXPY to carry out the experiments to test linear separability. The tool developed by Zhong et al. \cite{zhong2024} was not available at the time of writing; LIBSVM \cite{chang2011} did not align with our goal to test perfect separability -- confirming that a hyperplane is a valid separator is straightforward, establishing non-separability is difficult based on approximate outcomes.

\section{Linear Separability of the MNIST Dataset}\label{sec:results}

This section presents the results of linear separability tests performed on the MNIST dataset. We begin with providing details of the experimental environment, followed by subsections that summarize the outcomes of the pairwise and the one-vs-rest analysis.

\subsection{Experimental Environment}\label{sec:setup}

The experiments were carried out using Google Colaboratory (Colab) \cite{google-colab} hosted Jupyter Notebook environment. We selected hardware accelerator type: "T4 GPU", with CPU Intel(R) Xeon(R) CPU @ 2.00 GHz, 12.7 GB of RAM, and a T4 GPU.
We used CVXPY version 1.6.7\footnote{CVXPY added GPU solver support starting with version 1.7. Since version 1.6.7 was used, GPU acceleration was not utilized.} that was the latest version available in Python 3.11.13 (default Python version in Colab, at the time of writing).

We formulated the linear separability problem as a CVXPY linear program, which served as the core for all experiments (only the datasets were adjusted for the specific investigation), as shown below:
\vfill 

\begin{lstlisting}[language=Python,basicstyle=\small]
import cvxpy as cp
...
w = cp.Variable(number_of_features)
b = cp.Variable()

# constraints cs
cs = [cp.multiply(y, X @ w + b) >= 1] 

# problem p
p = cp.Problem(cp.Minimize(0), cs) 
p.solve()    
...
\end{lstlisting}

The decision variables are: \texttt{w} and \texttt{b}, and the objective is to minimize the constant 0, making this a pure feasibility problem (\texttt{p}). Matrix \texttt{X} contains all data samples (one per row), while vector \texttt{y} holds the corresponding labels (0--9). The \texttt{@} operator performs matrix-vector multiplication, forming constraints corresponding to equation (\ref{eqn2}). No solver was specified for \texttt{problem.solve()}, CVXPY automatically selected an appropriate (CLARABEL \cite{Clarabel_2024}). If the solver reached an \textit{optimal} solution (\texttt{p.status==cp.OPTIMAL}), \texttt{w} and \texttt{b} gave the normal vector and bias of a valid separating hyperplane; otherwise, the dataset is judged non-separable (\texttt{cp.INFEASIBLE} status).

\subsection{Pairwise Linear Separability}\label{sec:pairwise}

In this section, we examine the pairwise linear separability of the MNIST dataset. In each iteration, two digits are chosen and two subsets are created: one containing images of the first digit (positive samples) and the other containing images of the second digit (negative samples). These subsets are tested for linear separability in a total of 45 combinations. This procedure is performed on the training, test, and combined sets separately. 

Table \ref{tab1} shows the outcomes for the MNIST training set. Samples of seven digit pairs (2--3, 2--8, 3--5, 3--8, 4--9, 5--8, and 7--9) were found to be non-separable, whereas samples of three digits (0, 1, and 6) proved to be separable from all other digits in pairwise comparisons. Intuitively, it suggests that these latter three digits are probably the easiest to recognize and distinguish from any other digit, while digit 8 is the most challenging, as it conflicts with the most, three other digits (2, 3, and 5).

\begin{table*}
\caption{\textbf{Pairwise} linear separability of the MNIST \textbf{training set}}
\scriptsize
\setlength{\tabcolsep}{13.6pt}
\begin{tabular}{c*{9}{c}}
\hline

\makecell{\textbf{digit}\\(samples)} &\makecell{\textbf{1}\\(6742)} &\makecell{\textbf{2}\\(5958)} &\makecell{\textbf{3}\\(6131)} &\makecell{\textbf{4}\\(5842)} &\makecell{\textbf{5}\\(5421)} &\makecell{\textbf{6}\\(5918)} &\makecell{\textbf{7}\\(6265)} &\makecell{\textbf{8}\\(5851)} &\makecell{\textbf{9}\\(5949)} \\

\hline

\makecell{\textbf{0}\\(5923)} &\cellcolor{green!20}Yes &\cellcolor{green!20}Yes &\cellcolor{green!20}Yes &\cellcolor{green!20}Yes &\cellcolor{green!20}Yes &\cellcolor{green!20}Yes &\cellcolor{green!20}Yes &\cellcolor{green!20}Yes &\cellcolor{green!20}Yes \\
\makecell{\textbf{1}\\(6742)} &&\cellcolor{green!20}Yes &\cellcolor{green!20}Yes &\cellcolor{green!20}Yes &\cellcolor{green!20}Yes &\cellcolor{green!20}Yes &\cellcolor{green!20}Yes &\cellcolor{green!20}Yes &\cellcolor{green!20}Yes \\
\makecell{\textbf{2}\\(5958)}&&&\cellcolor{red!20}No  &\cellcolor{green!20}Yes &\cellcolor{green!20}Yes &\cellcolor{green!20}Yes &\cellcolor{green!20}Yes &\cellcolor{red!20}No  &\cellcolor{green!20}Yes \\
\makecell{\textbf{3}\\(6131)} &&&&\cellcolor{green!20}Yes &\cellcolor{red!20}No&\cellcolor{green!20}Yes &\cellcolor{green!20}Yes &\cellcolor{red!20}No  &\cellcolor{green!20}Yes \\
\makecell{\textbf{4}\\(5842)} &&&&&\cellcolor{green!20}Yes &\cellcolor{green!20}Yes &\cellcolor{green!20}Yes &\cellcolor{green!20}Yes &\cellcolor{red!20}No  \\
\makecell{\textbf{5}\\(5421)} &&&&&&\cellcolor{green!20}Yes &\cellcolor{green!20}Yes &\cellcolor{red!20}No  &\cellcolor{green!20}Yes \\
\makecell{\textbf{6}\\(5918)} &&&&&& &\cellcolor{green!20}Yes &\cellcolor{green!20}Yes &\cellcolor{green!20}Yes \\
\makecell{\textbf{7}\\(6265)} &&&&&&& &\cellcolor{green!20}Yes &\cellcolor{red!20}No  \\
\makecell{\textbf{8}\\(5851)} &&&&&&&& &\cellcolor{green!20}Yes \\

\end{tabular}
\label{tab1}
\end{table*}

\begin{table*}
\caption{\textbf{Pairwise} linear separability of of the combined MNIST \textbf{training and test set}}
\scriptsize
\setlength{\tabcolsep}{13.6pt}
\begin{tabular}{c*{9}{c}}
\hline

\makecell{\textbf{digit}\\(samples)}      &\makecell{\textbf{1}\\(7877)} &\makecell{\textbf{2}\\(6990)} &\makecell{\textbf{3}\\(7141)} &\makecell{\textbf{4}\\(6824)} &\makecell{\textbf{5}\\(6313)} &\makecell{\textbf{6}\\(6876)} &\makecell{\textbf{7}\\(7293)} &\makecell{\textbf{8}\\(6825)} &\makecell{\textbf{9}\\(6958)} \\

\hline

\makecell{\textbf{0}\\(6903)} &\cellcolor{green!20}Yes &\cellcolor{green!20}Yes &\cellcolor{green!20}Yes &\cellcolor{green!20}Yes &\cellcolor{green!20}Yes &\cellcolor{green!20}Yes &\cellcolor{green!20}Yes &\cellcolor{green!20}Yes &\cellcolor{green!20}Yes \\
\makecell{\textbf{1}\\(7877)} &&\cellcolor{green!20}Yes &\cellcolor{green!20}Yes &\cellcolor{green!20}Yes &\cellcolor{green!20}Yes &\cellcolor{green!20}Yes &\cellcolor{green!20}Yes &\cellcolor{green!20}Yes &\cellcolor{green!20}Yes \\
\makecell{\textbf{2}\\(6990)}&&&\cellcolor{red!20}No  &\cellcolor{green!20}Yes &\cellcolor{green!20}Yes &\cellcolor{green!20}Yes &\cellcolor{green!20}Yes &\cellcolor{red!20}No  &\cellcolor{green!20}Yes \\
\makecell{\textbf{3}\\(7141)} &&&&\cellcolor{green!20}Yes &\cellcolor{red!20}No&\cellcolor{green!20}Yes &\cellcolor{green!20}Yes &\cellcolor{red!20}No  &\cellcolor{green!20}Yes \\
\makecell{\textbf{4}\\(6824)} &&&&&\cellcolor{green!20}Yes &\cellcolor{green!20}Yes &\cellcolor{green!20}Yes &\cellcolor{green!20}Yes &\cellcolor{red!20}No  \\
\makecell{\textbf{5}\\(6313)} &&&&&&\cellcolor{green!20}Yes &\cellcolor{green!20}Yes &\cellcolor{red!20}No  &\cellcolor{green!20}Yes \\
\makecell{\textbf{6}\\(6876)} &&&&&& &\cellcolor{green!20}Yes &\cellcolor{green!20}Yes &\cellcolor{green!20}Yes \\
\makecell{\textbf{7}\\(7293)} &&&&&&& &\cellcolor{green!20}Yes &\cellcolor{red!20}No  \\
\makecell{\textbf{8}\\(6825)} &&&&&&&& &\cellcolor{green!20}Yes \\

\end{tabular}
\label{tab3}
\end{table*}

The results also support the results of Zhong et al. \cite{zhong2024}, who ranked all digit pairs by their \textit{linear separability (LS) degree}  (Fig. 29 in \cite{zhong2024}). Lower LS degree implies a smaller likelihood of finding the sets to be linearly separable. The first six pairs, as they listed: 3--5, 3--8, 4--9, 7--9, 5--8, 2--8, with the lowest LS degree, proved to be non-separable by our experiments, while other pairs: 2--6, 2--7, 8--9, etc., with higher LS degree, were found separable. The only outlier is pair 5--6, which was separable, yet had a slightly lower LS degree than the non-separable pair 2--3.

The execution times (as reported by CVXPY's \texttt{solver\_stats.solve\_time}) of testing linear separability of the individual digit pairs in the training set were in the range of 6.4--13.6 seconds for separable pairs and in the range of 15.9--24.7 seconds for the non-separable cases, as shown in table \ref{tab2} (first row in the corresponding cells). We note that the times of one representative execution are reported, with the aim to provide indicative values regarding the lengths of the different calculations; repeated experiments exhibited similar performance. 

Table \ref{tab3} shows the results on the entire MNIST dataset, combining both the training and the test samples. Notably, the addition of the test data does not alter the separability of any digit pairs observed previously (only sample counts differ). This indicates that, whenever a separating hyperplane exists, it is theoretically possible to find one that achieves perfect separation on the (unseen) test set -- teaching the model exclusively on the training set. 

The execution times in this case varied between 7.7 and 17.7 seconds for separable, and 18.6 and 28.9 seconds for non-separable digit pairs (see table \ref{tab2}, second row of the corresponding cells). 

Table \ref{tab5} reports the pairwise linear separability results for the test set. All digit pairs appear separable, which is due to the small sample size (as it does hold for the larger, training set). The execution times ranged from 0.6 to 2.0 seconds (see table \ref{tab2}, third row of the corresponding cells).

\begin{table*}
\caption{\textbf{Pairwise} linear separability of the MNIST \textbf{test set}}
\scriptsize
\setlength{\tabcolsep}{14.5pt}
\begin{tabular}{c*{9}{c}}
\hline

\makecell{\textbf{digit}\\(samples)}      &\makecell{\textbf{1}\\(1135} &\makecell{\textbf{2}\\(1032)} &\makecell{\textbf{3}\\(1010)} &\makecell{\textbf{4}\\(982)} &\makecell{\textbf{5}\\(892)} &\makecell{\textbf{6}\\(958)} &\makecell{\textbf{7}\\(1028)} &\makecell{\textbf{8}\\(974)} &\makecell{\textbf{9}\\(1009)} \\

\hline

\makecell{\textbf{0}\\(980)} &\cellcolor{green!20}Yes &\cellcolor{green!20}Yes &\cellcolor{green!20}Yes &\cellcolor{green!20}Yes &\cellcolor{green!20}Yes &\cellcolor{green!20}Yes &\cellcolor{green!20}Yes &\cellcolor{green!20}Yes &\cellcolor{green!20}Yes \\
\makecell{\textbf{1}\\(1135)} &&\cellcolor{green!20}Yes &\cellcolor{green!20}Yes &\cellcolor{green!20}Yes &\cellcolor{green!20}Yes &\cellcolor{green!20}Yes &\cellcolor{green!20}Yes &\cellcolor{green!20}Yes &\cellcolor{green!20}Yes \\
\makecell{\textbf{2}\\(1032)}&&&\cellcolor{green!20}Yes  &\cellcolor{green!20}Yes &\cellcolor{green!20}Yes &\cellcolor{green!20}Yes &\cellcolor{green!20}Yes &\cellcolor{green!20}Yes &\cellcolor{green!20}Yes \\
\makecell{\textbf{3}\\(1010)} &&&&\cellcolor{green!20}Yes &\cellcolor{green!20}Yes &\cellcolor{green!20}Yes &\cellcolor{green!20}Yes &\cellcolor{green!20}Yes  &\cellcolor{green!20}Yes \\
\makecell{\textbf{4}\\(982)} &&&&&\cellcolor{green!20}Yes &\cellcolor{green!20}Yes &\cellcolor{green!20}Yes &\cellcolor{green!20}Yes &\cellcolor{green!20}Yes \\
\makecell{\textbf{5}\\(892)} &&&&&&\cellcolor{green!20}Yes &\cellcolor{green!20}Yes &\cellcolor{green!20}Yes  &\cellcolor{green!20}Yes \\
\makecell{\textbf{6}\\(958)} &&&&&& &\cellcolor{green!20}Yes &\cellcolor{green!20}Yes &\cellcolor{green!20}Yes \\
\makecell{\textbf{7}\\(1028)} &&&&&&& &\cellcolor{green!20}Yes &\cellcolor{green!20}Yes  \\
\makecell{\textbf{8}\\(974)} &&&&&&&& &\cellcolor{green!20}Yes \\

\end{tabular}
\label{tab5}
\end{table*}

These latter separability outcomes also align with the findings of Zhong et al. (Table VI in \cite{zhong2024}). The execution times we measured, however, were shorter than their reported range of 4.95--7.91 seconds, on a similar CPU (Intel(R) Xeon(R) Silver 4210 2.20 GHz, in Matlab environment). For a direct comparison, the related execution times are collected in table \ref{tab7} (for digit pairs containing digit 0). We observed a 4--8x speedup compared to their method. We note that this comparison was on a limited dataset and measurements and cannot be considered generally conclusive.


\begin{table*}
\caption{Performance comparison of pairwise linear separability tests of digit 0 in the MNIST test set}
\footnotesize

\setlength{\tabcolsep}{9pt}
\begin{tabular}{c*{10}{c}}
\hline

\makecell{\textbf{pair}\\(all samples)} &\makecell{\textbf{0--1}\\(2115)} &\makecell{\textbf{0--2}\\(2012)} &\makecell{\textbf{0--3}\\(1990)} &\makecell{\textbf{0--4}\\(1962)} &\makecell{\textbf{0-5}\\(1872)} &\makecell{\textbf{0--6}\\(1938)} &\makecell{\textbf{0--7}\\(2008)} &\makecell{\textbf{0--8}\\(1985)} &\makecell{\textbf{0--9}\\(1989)}  \\

\hline

\makecell{Zhong's method \cite{zhong2024}} &4.95s &7.91s &7.18s &6.99s &6.05s &5.9s &7.44s &6.85s &6.9s \\
\makecell{CVXPY feasibility} &1.1s &2.0s &1.2s &1.1s &1.2s &1.2s &0.9s &1.5s &1.4s \\

\end{tabular}
\label{tab7}
\end{table*}

\begin{table}
\caption{\textbf{Execution times} of \textbf{pairwise} linear separability tests of the MNIST training (first row in the cell), combined (second row) and the test set (third row). (Boldface numbers denote \textit{non-separable} pairs.)}

\scriptsize

\setlength{\tabcolsep}{15pt}
\renewcommand{\arraystretch}{3}

\begin{tabular}{c*{9}{c}}

\hline

\makecell{\textbf{digit}\\execution time} &\makecell{\textbf{1}} &\makecell{\textbf{2}} &\makecell{\textbf{3}} &\makecell{\textbf{4}} &\makecell{\textbf{5}} &\makecell{\textbf{6}} &\makecell{\textbf{7}} &\makecell{\textbf{8}} &\makecell{\textbf{9}} \\

\hline

\makecell{\textbf{0}} &\makecell{12.3s\\9.9s\\1.1s} &\makecell{11.9s\\14.5s\\2.0s} &\makecell{10.8s\\13.6s\\1.2s} &\makecell{10.2s\\11.2s\\1.1s} &\makecell{10.8s\\12.4s\\1.2s} &\makecell{11.1s\\13.9s\\1.2s} &\makecell{11.1s\\12.8s\\0.9s} &\makecell{11.9s\\14.5s\\1.5s} &\makecell{10.9s\\12.2s\\1.4s} \\
\makecell{\textbf{1}} &&\makecell{11.2s\\14.3s\\1.1s} &\makecell{9.8s\\11.6s\\0.7s} &\makecell{6.8\\8.6\\0.8s} &\makecell{6.6s\\8.5s\\0.6s} &\makecell{9.5s\\10.9s\\0.9s} &\makecell{7.2s\\7.7s\\0.6s} &\makecell{13.0s\\16.6s\\0.7s} &\makecell{6.4s\\9.5s\\0.8s} \\
\makecell{\textbf{2}}&&&\makecell{\textbf{24.7s}\\\textbf{28.4s}\\1.1s} &\makecell{12.7s\\13.7s\\1.1s} &\makecell{13.0s\\16.1s\\1.1s} &\makecell{12.7s\\15.9s\\1.7s} &\makecell{13.6s\\17.3s\\1.1s} &\makecell{\textbf{22.6s}\\\textbf{27.4s}\\0.9s} &\makecell{13.6s\\16.1s\\0.9s} \\
\makecell{\textbf{3}} &&&&\makecell{10.4s\\13.2s\\1.2s} &\makecell{\textbf{23.5s}\\\textbf{23.7s}\\1.1s} &\makecell{11.9s\\13.0s\\1.1s} &\makecell{13.3s\\15.7s\\1.1s} &\makecell{\textbf{23.3s}\\\textbf{28.9s}\\1.2s} &\makecell{12.7s\\17.7s\\1.5s} \\
\makecell{\textbf{4}} &&&&&\makecell{9.7s\\12.0s\\1.1s} &\makecell{11.2s\\13.1s\\0.9s} &\makecell{9.6s\\13.0s\\0.8s} &\makecell{10.9s\\11.4s\\1.0s} &\makecell{\textbf{18.0s}\\\textbf{22.8s}\\1.0s} \\
\makecell{\textbf{5}} &&&&&&\makecell{12.3s\\17.6s\\1.0s} &\makecell{11.3s\\12.5s\\1.0s} &\makecell{\textbf{15.9s}\\\textbf{18.6s}\\1.1s} &\makecell{12.0s\\13.9s\\0.8s} \\
\makecell{\textbf{6}} &&&&&&&\makecell{10.9s\\12.1s\\1.3s} &\makecell{11.5s\\14.0s\\1.3s} &\makecell{11.4s\\12.4s\\1.0s} \\
\makecell{\textbf{7}} &&&&&&&&\makecell{12.2s\\15.0s\\0.8s} &\makecell{\textbf{17.8s}\\\textbf{23.5s}\\0.8s} \\
\makecell{\textbf{8}} &&&&&&&&&\makecell{12.3s\\17.2s\\1.0s} \\

\end{tabular}
\label{tab2}
\end{table}

\subsection{One-vs-rest Linear Separability}\label{sec:one-vs-rest}

In this subsection, we report the one-vs-rest linear separability results of the MNIST dataset. The goal is to determine whether images of a given digit (positive class) can be linearly separated from images of all other digits (negative class). It requires investigating each of the 10 digits, in datasets training, test and the combined one.

Table \ref{tab8} summarizes the results obtained on the training set. The outcomes show that, regardless of which digit is chosen, the samples cannot be linearly separated from the images of the other digits. The execution times, in this case, varied between 89 and 209 seconds due to the larger set sizes, as shown in table \ref{tab11} (first row).


\begin{table*}[t]
\caption{\textbf{One-vs-rest} linear separability of the MNIST \textbf{training set}}
\scriptsize

\setlength{\tabcolsep}{7.6pt}
\begin{tabular}{c*{10}{c}}
\hline

\makecell{\textbf{digit}\\(positive samples)} &\makecell{\textbf{0}\\(5923)} &\makecell{\textbf{1}\\(6742)} &\makecell{\textbf{2}\\(5958)} &\makecell{\textbf{3}\\(6131)} &\makecell{\textbf{4}\\(5842)} &\makecell{\textbf{5}\\(5421)} &\makecell{\textbf{6}\\(5918)} &\makecell{\textbf{7}\\(6265)} &\makecell{\textbf{8}\\(5851)} &\makecell{\textbf{9}\\(5949)} \\

\hline

\makecell{\textbf{1-vs-rest} linear separability\\(negative samples)} &\cellcolor{red!20}\makecell{No\\(54077)} &\cellcolor{red!20}\makecell{No\\(53258)} &\cellcolor{red!20}\makecell{No\\(54042)} &\cellcolor{red!20}\makecell{No\\(53869)} &\cellcolor{red!20}\makecell{No\\(54158)} &\cellcolor{red!20}\makecell{No\\(54579)} &\cellcolor{red!20}\makecell{No\\(54082)} &\cellcolor{red!20}\makecell{No\\(53735)} &\cellcolor{red!20}\makecell{No\\(54149)} &\cellcolor{red!20}\makecell{No\\(54051)}\\
\end{tabular}
\label{tab8}
\end{table*}

Most of these findings were expected, since as indicated in table \ref{tab1}, digits: 2, 3, 4, 5, 7, 8, and 9 were already known to be non-separable in pairwise comparisons. Only digits 0, 1, and 6 remained uncertain, which are now also proved to be non-separable from the rest. 

Table \ref{tab9} presents the results of the combined set. It is only included for completeness, since training-set results (table \ref{tab8}) already anticipated these outcomes -- adding new data to non-separable sets cannot make them separable. The execution times of testing one-vs-rest separability ranged from 114 to 189 seconds in the combined set, as shown in table \ref{tab11} (second row).

\begin{table*}
\caption{\textbf{One-vs-rest} linear separability of the combined MNIST \textbf{training and test set}}
\scriptsize
\setlength{\tabcolsep}{7.6pt}
\begin{tabular}{c*{10}{c}}
\hline

\makecell{\textbf{digit}\\(positive samples)} 
&\makecell{\textbf{0}\\(6903)} &\makecell{\textbf{1}\\(7877)} &\makecell{\textbf{2}\\(6990)} &\makecell{\textbf{3}\\(7141)} &\makecell{\textbf{4}\\(6824)} &\makecell{\textbf{5}\\(6313)} &\makecell{\textbf{6}\\(6876)} &\makecell{\textbf{7}\\(7293)} &\makecell{\textbf{8}\\(6825)} &\makecell{\textbf{9}\\(6958)} \\

\hline
\makecell{\textbf{1-vs-rest} linear separability\\(negative samples)} 

&\cellcolor{red!20}\makecell{No\\(63097)} 
&\cellcolor{red!20}\makecell{No\\(62123)} &\cellcolor{red!20}\makecell{No\\(63010)} &\cellcolor{red!20}\makecell{No\\(62859)} &\cellcolor{red!20}\makecell{No\\(63176)} &\cellcolor{red!20}\makecell{No\\(63687)} &\cellcolor{red!20}\makecell{No\\(63124)} &\cellcolor{red!20}\makecell{No\\(62707)} &\cellcolor{red!20}\makecell{No\\(63175)} &\cellcolor{red!20}\makecell{No\\(63042)}\\
\end{tabular}
\label{tab9}
\end{table*}

\begin{table*}[!htbp] 
\caption{\textbf{One-vs-rest} linear separability of the MNIST \textbf{test set}}
\scriptsize

\setlength{\tabcolsep}{9pt}
\begin{tabular}{c*{10}{c}}
\hline

\makecell{\textbf{digit}\\(positive samples)} 
&\makecell{\textbf{0}\\(980)} &\makecell{\textbf{1}\\(1135)} &\makecell{\textbf{2}\\(1032)} &\makecell{\textbf{3}\\(1010)} &\makecell{\textbf{4}\\(982)} &\makecell{\textbf{5}\\(892)} &\makecell{\textbf{6}\\(958)} &\makecell{\textbf{7}\\(1028)} &\makecell{\textbf{8}\\(974)} &\makecell{\textbf{9}\\(1009)} \\

\hline

\makecell{\textbf{1-vs-rest} linear separability\\(negative samples)} 
&\cellcolor{green!20}\makecell{Yes\\(9020)} &\cellcolor{green!20}\makecell{Yes\\(8865)} &\cellcolor{green!20}\makecell{Yes\\(8968)} &\cellcolor{green!20}\makecell{Yes\\(8990)} &\cellcolor{green!20}\makecell{Yes\\(9018)} &\cellcolor{red!20}\makecell{No\\(9108)} &\cellcolor{green!20}\makecell{Yes\\(9042)} &\cellcolor{green!20}\makecell{Yes\\(8972)} &\cellcolor{red!20}\makecell{No\\(9026)} &\cellcolor{red!20}\makecell{No\\(8991)}\\

\end{tabular}
\label{tab10}
\end{table*}

Finally, table \ref{tab10} shows the results obtained on the test set. Several digits (0--4, 6 and 7) proved to be linearly separable from all others. However, given the small sample size (and earlier results), these findings are not conclusive with respect to the one-vs-rest separability of these digits. The execution times fall in the range of 9.5--22.6 seconds, as shown in table \ref{tab11} (third row).

\begin{table}
\caption{\textbf{Execution times} of the \textbf{one-vs-rest} linear separability tests of the MNIST training, combined (training+test), and the test sets. (Boldface numbers denote \textit{non-separable} cases.)}

\scriptsize

\setlength{\tabcolsep}{12.5pt}
\begin{tabular}{c*{10}{c}}
\hline
\makecell{\textbf{digit}\\execution time}&\makecell{\textbf{0}}&\makecell{\textbf{1}}&\makecell{\textbf{2}}&\makecell{\textbf{3}}&\makecell{\textbf{4}}&\makecell{\textbf{5}}&\makecell{\textbf{6}}&\makecell{\textbf{7}}&\makecell{\textbf{8}}&\makecell{\textbf{9}}\\
\hline
\makecell{\textbf{training set}} &\textbf{130s} &\textbf{209s} &\textbf{95s} &\textbf{122s} &\textbf{114s} &\textbf{89s} &\textbf{140s} &\textbf{128s} &\textbf{99s} &\textbf{112s} \\
\makecell{\textbf{training+test}} &\textbf{158s} &\textbf{189s} &\textbf{115s} &\textbf{148s} &\textbf{139s} &\textbf{114s} &\textbf{155s} &\textbf{179s} &\textbf{116s} &\textbf{143s} \\
\makecell{\textbf{test set}} &9.5s &10.1s &20.8s &19.3s &10.2s &\textbf{22.6s} &11.3s &11.2s &\textbf{17.5s} &\textbf{21.5s} \\
\end{tabular}
\label{tab11}
\end{table}

\section{Conclusions}\label{sec:Conclusions}

This paper aimed at clarifying the question of linear separability of the MNIST dataset. The investigations extended to the training set, the test set, and their combination, respectively, examining linear separability in both the pairwise and one-vs-rest relations.

Based on the results presented, we can now judge the oversimplifying claims that often come up in formal and informal sources, such as "the handwritten dataset in MNIST is linearly separable" \cite{froch2025} or "MNIST is not linearly separable" \cite{perrotta2020programming}. Only the test set can definitely be declared as linearly separable -- but only in the pairwise case, whereas the training set (and so the entire dataset) can be declared as non-separable -- but only in the one-vs-rest case. All other cases yield mixed outcomes.

The applied method used the CVXPY convex optimization library, which showed outstanding efficiency, though the comparison with other methods was limited due to the scarcity of relevant data. The reported performance results can serve as a baseline for future studies and improvements.

\section*{Dataset} The MNIST dataset is accessible from within the TensorFlow-Keras framework (\texttt{tensorflow.keras.datasets.mnist}) -- other sources are also available. For reproducibility of the results, the source codes of the experiments can be found in a  \href{https://github.com/ahajnal/MNIST-linear-separability}{GitHub repository}. 

\section*{Declaration interests}
The author declare that they have no known competing financial interests or personal relationships that could have appeared to influence the work reported in this paper.

\vfill 

\bibliographystyle{IEEEtran}
\bibliography{references}  
 
\vfill
\pagebreak

\end{document}